\documentclass[conference]{IEEEtran}
\usepackage{cite}
\usepackage{amsmath,amssymb,amsfonts}
\usepackage{algorithmic}
\usepackage{graphicx}
\usepackage{textcomp}
\usepackage{xcolor}
\usepackage{subcaption}
\usepackage{booktabs}
\usepackage{tabularx}
\usepackage{hyperref}
\newcolumntype{Y}{>{\centering\arraybackslash}X}

\def\BibTeX{{\rm B\kern-.05em{\sc i\kern-.025em b}\kern-.08em
    T\kern-.1667em\lower.7ex\hbox{E}\kern-.125emX}}
\begin{document}

\title{Directional Antenna Systems for Long-Range Through-Wall Human Activity Recognition
\thanks{Identify applicable funding agency here. If none, delete this.}
}

\author{\IEEEauthorblockN{Julian Strohmayer and Martin Kampel}
\IEEEauthorblockA{\textit{Computer Vision Lab, TU Wien} \\
Favoritenstr. 9/193-1, 1040 Vienna, Austria \\
\{julian.strohmayer, martin.kampel\}@tuwien.ac.at}
}

\maketitle

\begin{abstract}
WiFi Channel State Information (CSI)-based human activity recognition (HAR) enables contactless, long-range sensing in spatially constrained environments while preserving visual privacy. However, despite the presence of numerous WiFi-enabled devices around us, few expose CSI to users, resulting in a lack of sensing hardware options. Variants of the Espressif ESP32 have emerged as potential low-cost and easy-to-deploy solutions for WiFi CSI-based HAR. In this work, four ESP32-S3-based 2.4GHz directional antenna systems are evaluated for their ability to facilitate long-range through-wall HAR. Two promising systems are proposed, one of which combines the ESP32-S3 with a directional biquad antenna. This combination represents, to the best of our knowledge, the first demonstration of such a system in WiFi-based HAR. The second system relies on the built-in printed inverted-F antenna (PIFA) of the ESP32-S3 and achieves directionality through a plane reflector. In a comprehensive evaluation of line-of-sight (LOS) and non-line-of-sight (NLOS) HAR performance, both systems are deployed in an office environment spanning a distance of 18 meters across five rooms. In this experimental setup, the Wallhack1.8k dataset, comprising 1806 CSI amplitude spectrograms of human activities, is collected and made publicly available. Based on Wallhack1.8k, we train activity recognition models using the EfficientNetV2 architecture to assess system performance in LOS and NLOS scenarios. For the core NLOS activity recognition problem, the biquad antenna and PIFA-based systems achieve accuracies of 92.0$\pm$3.5 and 86.8$\pm$4.7, respectively, demonstrating the feasibility of long-range through-wall HAR with the proposed systems.
\end{abstract}


\section{Introduction}
\label{sec:introduction}
\label{sec:intro}


In indoor spaces, WiFi signal propagation is determined by the environment \cite{Lee10185958}. While static objects such as walls and furniture primarily contribute to the background signal, dynamic objects, such as humans, rapidly alter signal paths, generating characteristic CSI patterns that facilitate Human Activity Recognition (HAR) applications \cite{Liu87946434353}. Although camera-based methodologies currently dominate the HAR field, WiFi is steadily gaining recognition as a viable sensing modality. WiFi offers a multitude of advantages, including cost-effectiveness, unobtrusiveness, immunity to changes in illumination, and the protection of visual privacy by not capturing color or texture information – a crucial requirement in privacy-sensitive applications \cite{Arning2015}. Moreover, WiFi signals possess the capability to penetrate walls, thus facilitating contactless long-range activity sensing within spatially constrained environments, with operational ranges extending up to 35 meters indoors \cite{Zafari2019ASO}. This not only presents an economic advantage when compared to camera-based approaches that necessitate per-room deployment but also unlocks innovative possibilities, such as through-wall HAR, constituting the central focus of this work.

While early WiFi-based HAR approaches relied on the Received Signal Strength Indicator (RSSI), measuring the signal strength of the WiFi channel at the receiver \cite{Youssef3434859348}, most contemporary approaches are based on Channel State Information (CSI). CSI captures both the amplitude and phase information of WiFi channel subcarriers, endowing it with higher information density, which, in turn, allows for the recognition of finer-grained activities and enhances robustness against environmental effects \cite{Parameswaran2009IsRA}. Despite the fact that most WiFi devices inherently process CSI, few off-the-shelf devices give end-users access to this information. Consequently, CSI capture is only feasible through specific combinations of hardware and software. Examples of such configurations are the Intel NIC 5300 in conjunction with the Linux 802.11n CSI Tool \cite{LinuxCSITool} and various Atheros NIC variants (AR9580, AR9590, AR9344, and QCA9558) employing the Atheros CSI Tool \cite{AtherosCSITool}. Recent developments have expanded the accessibility of CSI capture to new platforms, such as the Raspberry Pi, utilizing the Nexmon CSI Tool \cite{NexmonCSITool}. Another emerging alternative is the Wi-ESP CSI Tool \cite{Wi-ESPCSITool}, which capitalizes on the popular ESP32 microcontroller manufactured by Espressif Systems. Although some works have explored the potential of the ESP32 in short-range line-of-sight (LOS) scenarios \cite{Hao2022} and non-line-of-sight (NLOS) HAR scenarios \cite{Hernandez2021, Kumar2022}, its activity sensing capabilities in long-range through-wall scenarios have remained unexplored.

\begin{figure*}[t!]
  \centering
  \begin{subfigure}{0.125\linewidth}
    \includegraphics[width=1.0\linewidth]{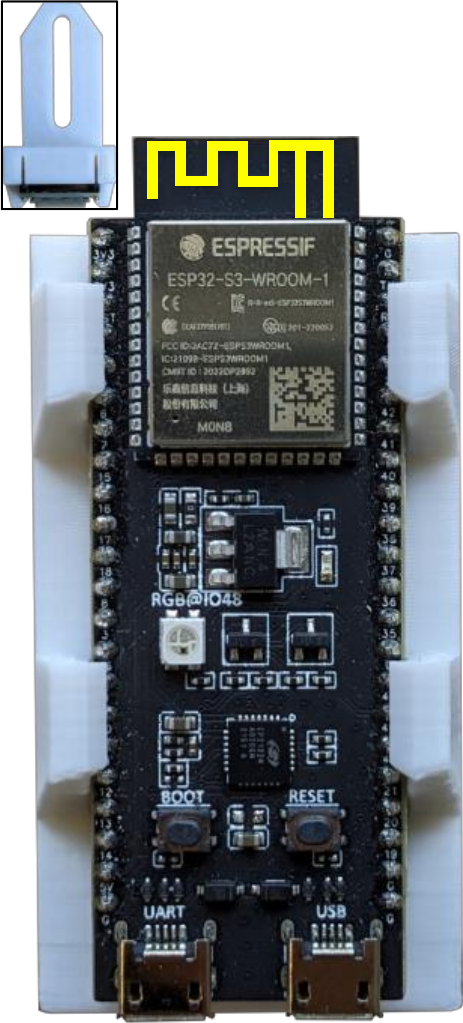}
    \caption{PIFA}
    \label{fig:antennasPCB}
  \end{subfigure}
\hfill
  \begin{subfigure}{0.24\linewidth}
    \includegraphics[width=1.0\linewidth]{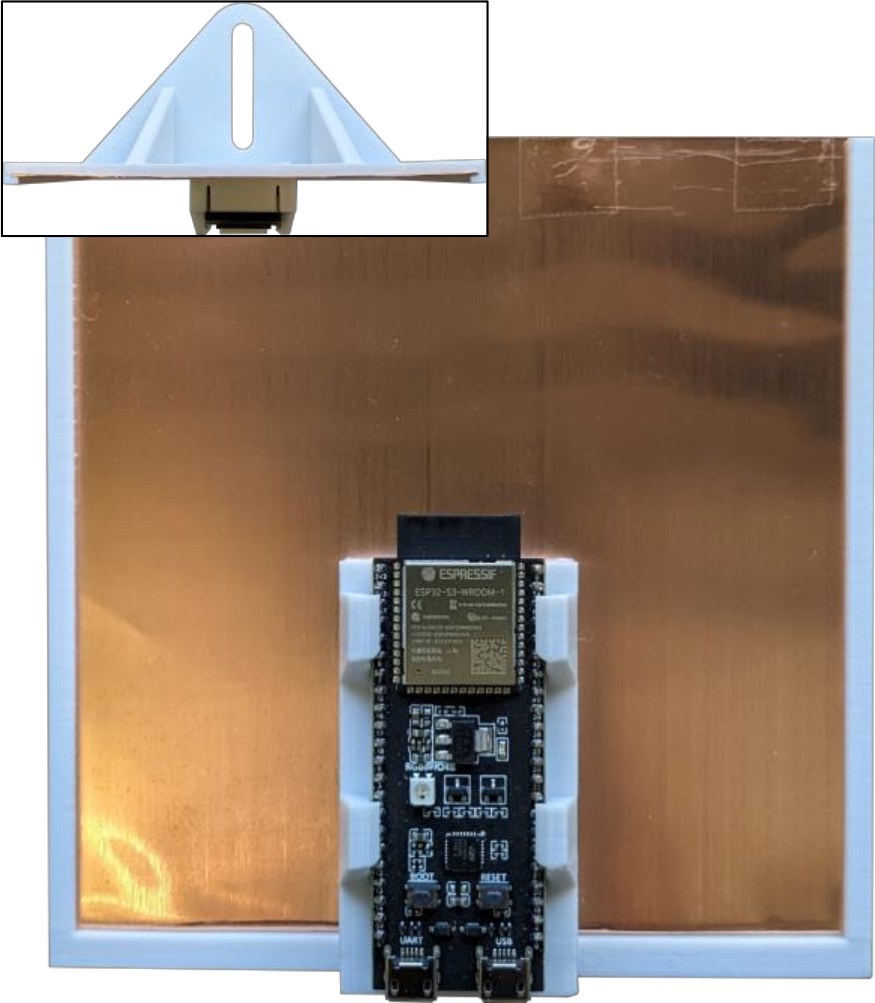}
    \caption{PIFA with plane reflector}
    \label{fig:antennasPCBPlane}
  \end{subfigure}
  \hfill
  \begin{subfigure}{0.355\linewidth}
    \includegraphics[width=1.0\linewidth]{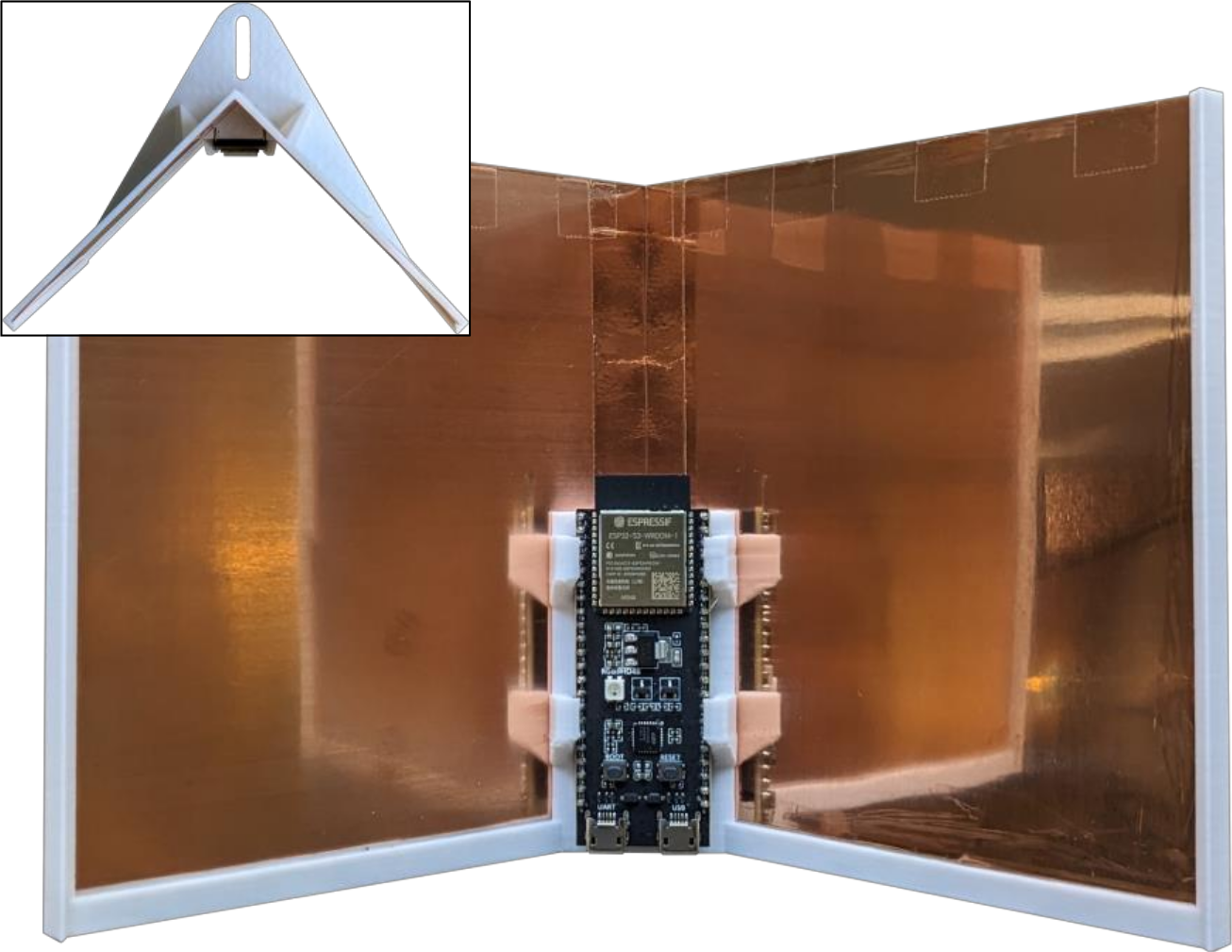}
    \caption{PIFA with 90$^{\circ}$ corner reflector}
    \label{fig:antennasPCBCorner}
  \end{subfigure}
  \hfill
  \begin{subfigure}{0.26\linewidth}
    \includegraphics[width=1.0\linewidth]{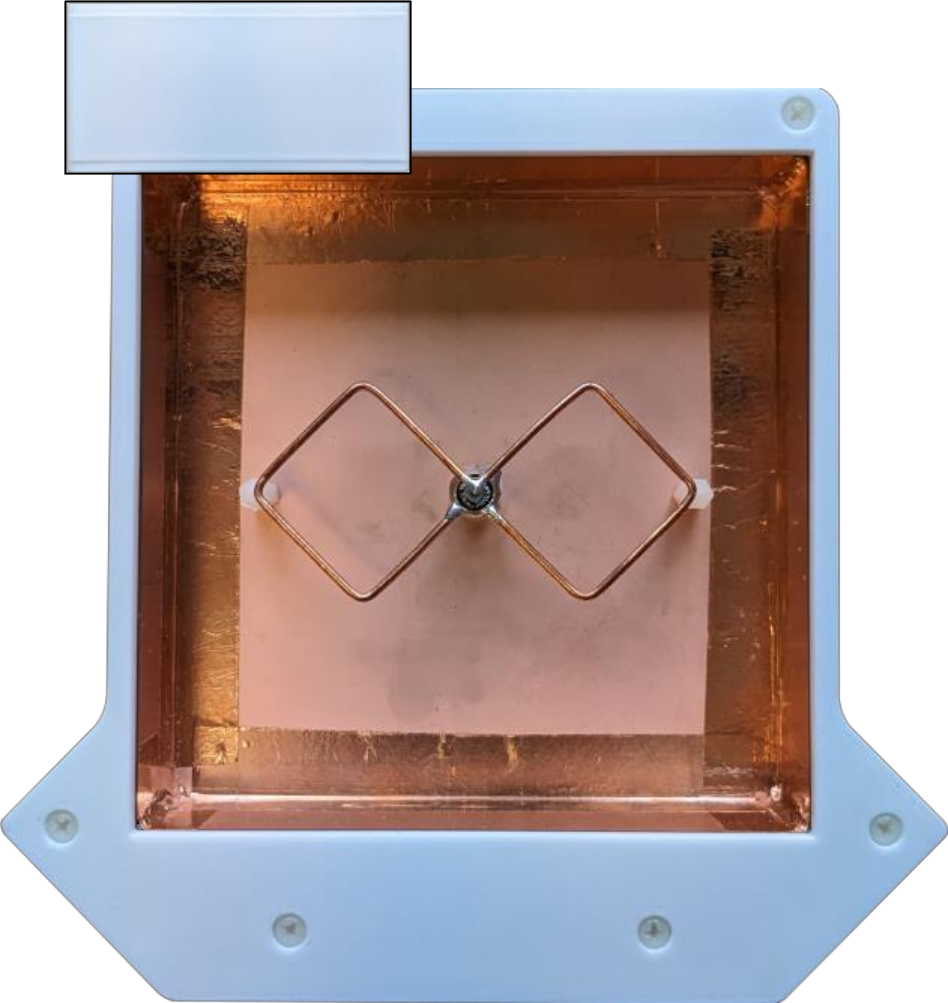}
    \caption{biquad antenna}
    \label{fig:antennasBiquad}
  \end{subfigure}
  \caption{Overview of the evaluated systems, showing  (a) the baseline system relying solely on the ESP32-S3's built-in PIFA, (b) PIFA with a plane reflector, (c) PIFA with a corner reflector, and (d) the external biquad antenna system.}
  \vspace{-5mm}
  \label{fig:systemsOverview}
\end{figure*}

\section{Related Work}
\label{sec:relatedWork}

Through-wall CSI-based HAR using the Intel NIC 5300 has been a topic of interest, as highlighted in the comprehensive survey by Wang et al. \cite{Wang2019}. However, with the discontinuation of the Intel NIC 5300 in 2016, its suitability for future CSI-based HAR applications is limited. As a result, researchers have explored alternatives, including the Espressif ESP32, which offers a cost-effective and easy-to-deploy solution. While the ESP32 has been utilized in LOS scenarios \cite{Atif2020, Hernandez2020, Hao2022}, NLOS scenarios have seen limited investigation. To the best of our knowledge, only two works have explored NLOS scenarios \cite{Hernandez2021, Kumar2022}.

The feasibility of adversarial occupancy monitoring based on CSI is assessed in the work by Hernandez and Bulut \cite{Hernandez2021}. ESP32 devices are positioned as transmitters and receivers on the external wall of a hallway, successfully sensing the presence and walking direction of humans. An interesting aspect of this work is the use of aluminum plates to act as RF shielding, enabling the side-by-side arrangement of the transmitter and receiver on the same wall while also enhancing signal strength by directing the built-in antenna of the ESP32. While our focus is on a conventional transmitter-receiver arrangement, with activities taking place between the devices, we draw inspiration from this work to explore the effects of antenna directionality in the context of long-range through-wall HAR scenarios.

Furthermore, Kumar et al. \cite{Kumar2022} employ ESP32-based systems to investigate presence and fall detection in NLOS scenarios. The systems are deployed in a conventional transmitter-receiver configuration. Experimental results show characteristic CSI patterns induced by activities, even with up to two walls between the transmitter and receiver. While these results hold promise, the limited evaluation, absence of key measurements such as transmitter-receiver spacing, and insufficient description of the recording environment hinder drawing a clear conclusion about the feasibility of long-range through-wall HAR using ESP32-based systems. Additionally, the proposed system utilizes an external low-gain omnidirectional rod antenna. This choice is not only inefficient due to a significant portion of emitted energy not being directed at the target area but also renders the system susceptible to noise from outside the recording environment. Building on the findings in \cite{Hernandez2021, Kumar2022}, promising approaches to long-range through-wall HAR with the ESP32 could encompass the use of RF shielding (reflectors) to eliminate noise and enforce directionality of the built-in antenna, or the integration of an external directional antenna – both of which are investigated in this work.

\section{Experimental Setup}
\label{sec:experiment}
In this section, we detail the experimental setup, encompassing the hardware components for all systems, the physical environment for LOS and NLOS performance evaluations, and the protocol for collecting CSI activity spectrograms used in training CNN-based regression models.

\subsection{Hardware}
\label{subsec:hardware}
We consider the four systems shown in Figure \ref{fig:systemsOverview}, all of which are built upon the ESP32-S3-DevKitC-1\footnote{ESP32-S3-DevKit-1, \href{https://docs.espressif.com/projects/esp-idf/en/latest/esp32s3/hw-reference/esp32s3/user-guide-devkitc-1.html}{https://docs.espressif.com}, accessed: 10-16-2023} development board featuring an ESP32-S3-WROOM-1\footnote{ESP32-S3-WROOM-1, \href{https://www.espressif.com/sites/default/files/documentation/esp32-s3-wroom-1_wroom-1u_datasheet_en.pdf}{https://docs.espressif.com}, accessed: 10-16-2023} module for WiFi connectivity. The systems are deployed in a symmetric transmitter-receiver configuration, where one of the two identical devices serves as a transmitter, sending CSI packets at a fixed frequency of 100Hz, while the other device functions as a receiver, continually listening for CSI packets. A WiFi connection between the transmitter and receiver is established using Espressif's wireless communication protocol ESP-NOW\footnote{ESP-NOW, \href{https://www.espressif.com/en/solutions/low-power-solutions/esp-now}{https://docs.espressif.com}, accessed: 10-16-2023}, and CSI packets are captured using Espressif's IoT Development Framework ESP-IDF\footnote{ESP-IDF, \href{https://docs.espressif.com/projects/esp-idf/en/latest/esp32/get-started/}{https://docs.espressif.com}, accessed: 10-16-2023}. Although all systems are based on the ESP32-S3-DevKitC-1, we can differentiate them based on the type of antenna employed. The systems depicted in Figure \ref{fig:antennasPCB}-\ref{fig:antennasPCBCorner} utilize the built-in antenna of the ESP32-S3-WROOM-1 module, whereas the system in Figure \ref{fig:antennasBiquad} replaces the built-in antenna with an external one.

\textbf{PIFA.}
Our baseline system, shown in Figure \ref{fig:antennasPCB}, is the unmodified ESP32-S3-DevKitC-1 development board that uses the built-in meandered printed inverted-F antenna (PIFA)\footnote{PIFA, \href{https://www.ti.com/lit/an/swra117d/swra117d.pdf}{https://www.ti.com}, accessed: 10-16-2023} \cite{pradhan2013parametric} of the ESP32-S3-WROOM-1 module. The PIFA can provide basic WiFi connectivity in most traditional scenarios; however, it is not ideal for long-range through-wall HAR applications. Its omnidirectionality not only prevents the constraining of the recording environment but also renders it susceptible to noise from outside the recording environment (e.g., a person walking behind the system or on the floor below) \cite{Hernandez2021}. Moreover, its low gain of 2dBi could hinder the establishment of a stable connection in long-range through-wall HAR scenarios.

\begin{figure}[h!]
  \centering
  \begin{subfigure}{0.5125\linewidth}
    \includegraphics[width=1.0\linewidth]{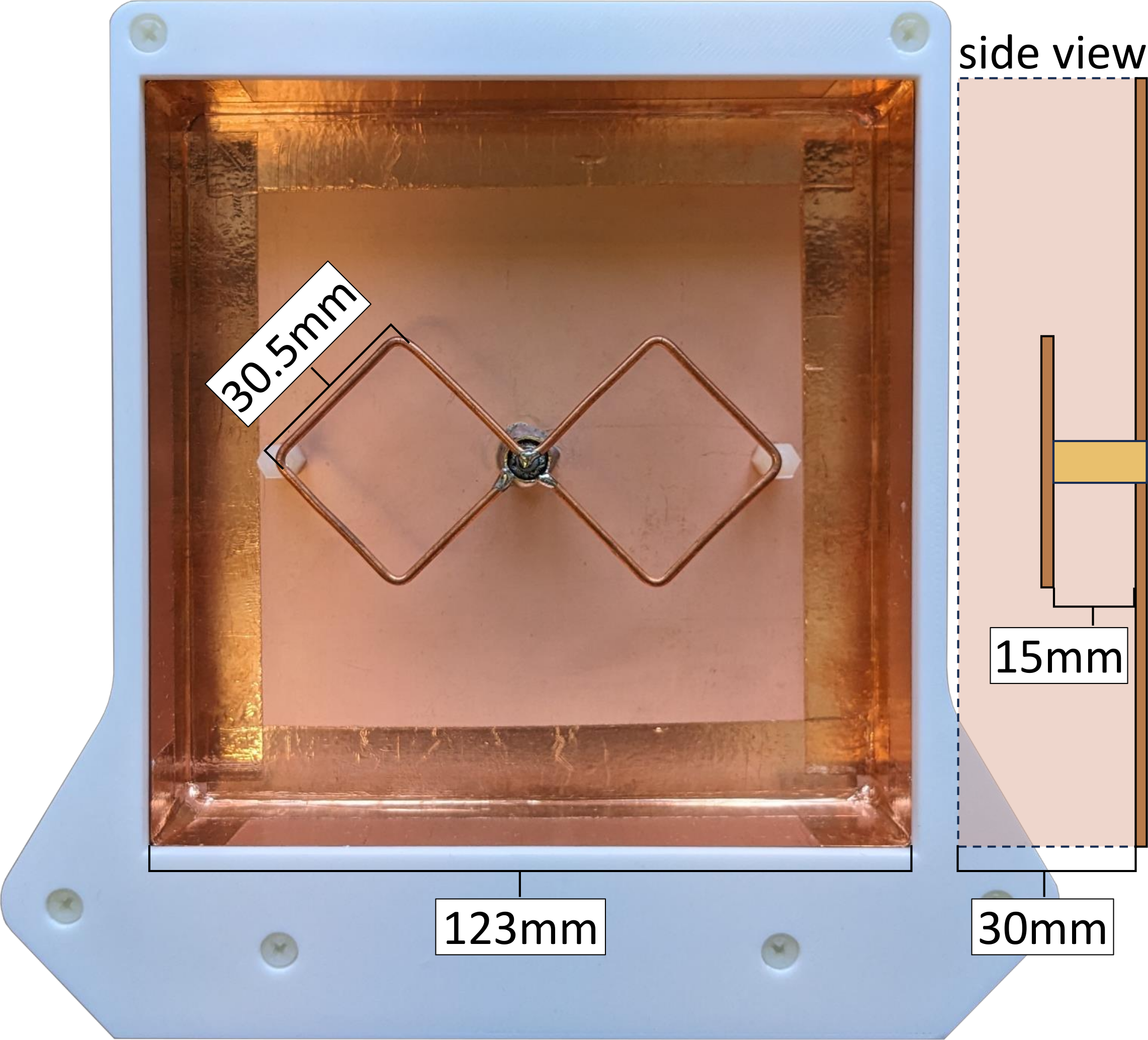}
    \caption{frontside}
    \label{fig:systemFront}
  \end{subfigure}
\hfill
  \begin{subfigure}{0.475\linewidth}
    \includegraphics[width=1.0\linewidth]{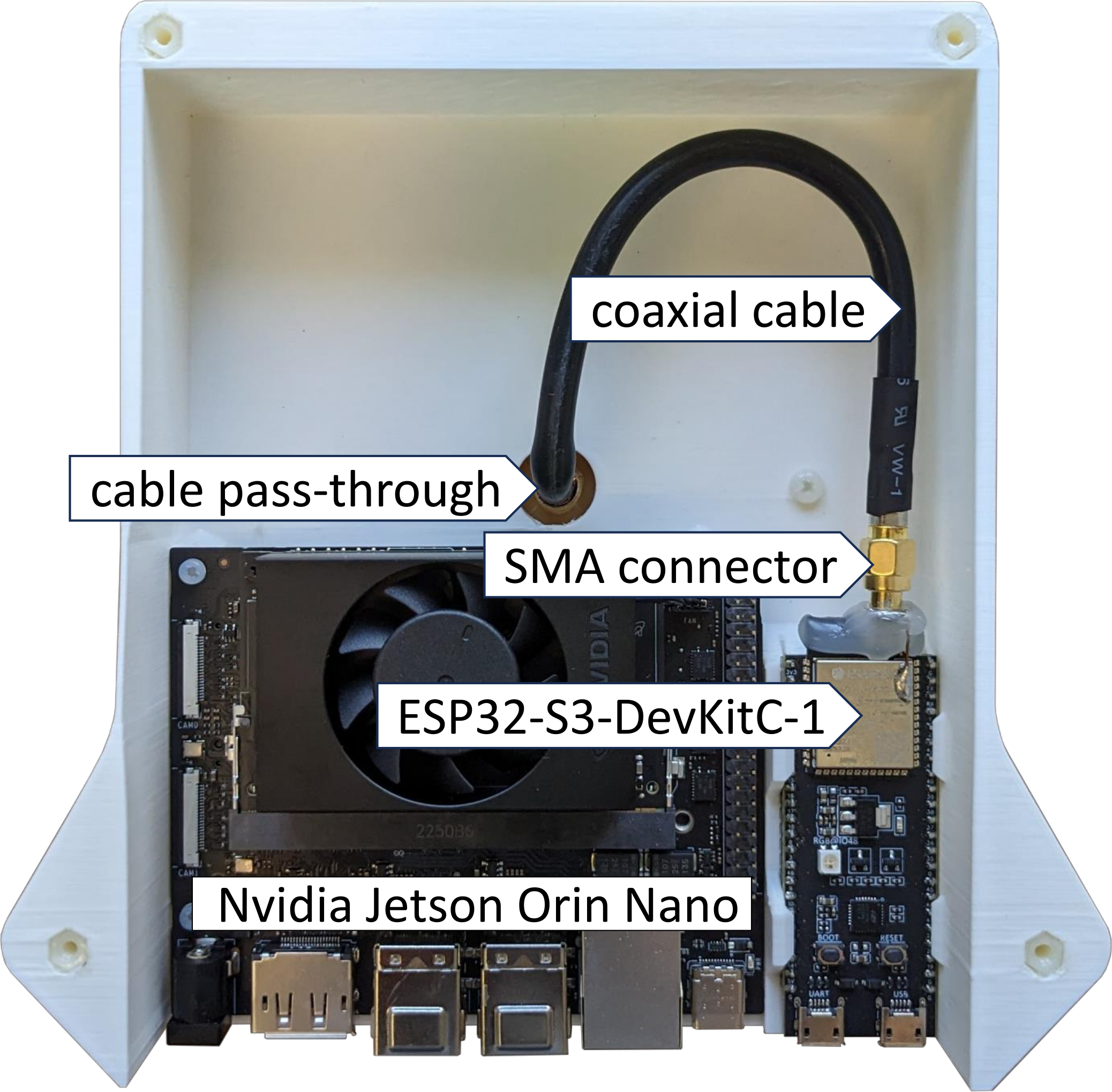}
    \caption{backside}
    \label{fig:systemBack}
  \end{subfigure}
  \caption{Overview of the proposed biquad antenna system, showing (a) antenna geometry in the frontal view, and (b) internal electronic components of the receiver unit.}
  \vspace{-6mm}
  \label{fig:system}
\end{figure}

\textbf{PIFA with plane reflector.}
To address these shortcomings without replacing the PIFA, we investigate the effects of using different reflector geometries on system performance. The first reflector-based system, shown in Figure \ref{fig:antennasPCBPlane}, uses a plane reflector made from a 123$\times$123mm, 0.2mm thick copper sheet. Both the ESP32-S3-DevKitC-1 development board and the reflector are rigidly mounted to a 3D printed frame, creating a 1/8-wavelength spacing of 15mm between the PIFA and the reflector. The added plane reflector eliminates noise originating from the backside of the PIFA and simultaneously increases its forward gain.

\textbf{PIFA with 90$^{\circ}$ corner reflector.}
Building on this idea, a second reflector-based system, shown in Figure \ref{fig:antennasPCBCorner}, is evaluated, which further narrows the beamwidth of the PIFA through the use of a 90$^{\circ}$ corner reflector. The reflector is constructed from two 123$\times$123mm, 0.2mm thick copper sheets joined by copper tape and held at a 90$^{\circ}$ angle using a 3D printed frame. Like the plane reflector system, a 1/8-wavelength spacing of 15mm is maintained between the corner of the reflector and the PIFA.

\textbf{Biquad antenna.}
Lastly, the fourth system in our evaluation, shown in Figure \ref{fig:antennasBiquad}, replaces the PIFA with an external antenna. We choose a directional biquad antenna design with a gain of 10-12dBi and a beamwidth of 70$^{\circ}$ \cite{Singh2012ANB}. This choice strikes a balance between gain, compactness, and ease of construction using common materials. Moreover, while antennas with higher gain and extremely narrow beam widths exist for establishing long-range point-to-point connections, we favor a beamwidth around 70$^{\circ}$ for HAR applications. Such a beamwidth facilitates comprehensive room coverage in most scenarios while maintaining constraints on the recording environment. Additionally, its similarity to the field of view (FOV) of typical cameras allows easy integration with a camera having a corresponding FOV, providing a sense of the antenna beam's coverage area.

For the antenna's construction, readily available materials are employed. The reflector, composed of the backplane and side lips, is constructed from blank copper PCB material with a single-sided 35$\mu$m copper layer. Detailed measurements of the antenna's geometry are given in Figure \ref{fig:systemFront}. The reflector's backplane measures 123$\times$123mm, and the side lips have a depth of 30mm. While the side lips could be omitted, their inclusion is beneficial as they reduce side-lobe power and enhance the antenna gain by 2dBi compared to a design without them \cite{Singh2012ANB}. Additionally, the side lips shield the radiating element from noise originating from sources orthogonal to the antenna's viewing direction. The radiating element uses a vertically polarized biquad geometry with a 1/4-wavelength edge length of 30.5mm, constructed from 2.5mm$^{2}$ solid core copper wire. This radiating element is mounted to a soldered copper tube, passing through the center of the reflector's backplane. To maintain a 1/8-wavelength spacing of 15mm between the radiating element and the reflector's backplane, two 15mm nylon standoffs are utilized. Lastly, as shown in Figure \ref{fig:systemBack}, a short segment of 50$\Omega$ impedance low-loss coaxial cable is soldered to the center of the radiating element and fed through the copper tube. The opposing end of the cable connects to the ESP32-S3-WROOM-1 module via an SMA connector, soldered to the signal and ground traces of the PIFA. For the sake of replication, CAD models of all 3D printed components are made publicly available\footnote{System CAD models, \href{https://zenodo.org/record/8188999}{https://zenodo.org} (available from 4-21-2024)}.

\begin{figure}[t!]
  \centering
  \includegraphics[width=\linewidth]{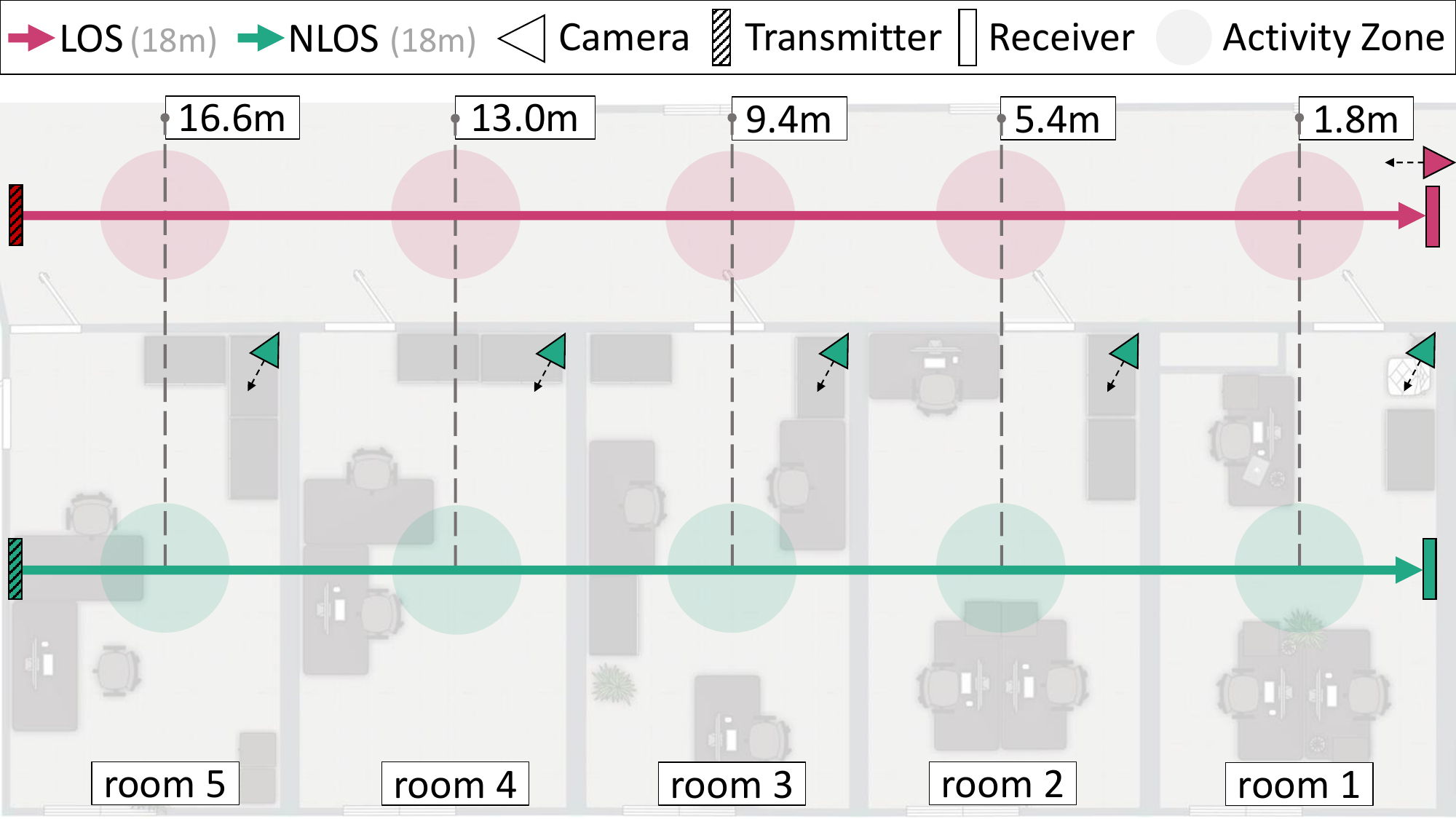}
  \caption{Floor plan of the evaluation environment, showing the transmitter and receiver placement in LOS and NLOS scenarios.}
  \label{environment}
  \vspace{-6mm}
\end{figure}

\vspace{-1mm}
\subsection{Environment}
\vspace{-1mm}
\label{subsec:environment}
The proposed systems are evaluated in the office environment depicted in Figure \ref{environment}. This environment comprises an 18m-long hallway connected to five adjacent rooms containing office furniture. These rooms, separated by 25cm thick brick walls, present a challenging long-range NLOS scenario. Moreover, the rooms are of uniform size (approximately 3.5m$\times$6.0m) and arranged in a manner that facilitates a direct comparison of LOS and NLOS HAR performance at various distances between the transmitter and receiver. For the LOS scenario (red line), the transmitter and receiver are positioned at opposite ends of the hallway, facing each other. To capture activity images required for the annotation of raw CSI data, an additional ESP32-S3-based camera board is placed next to the transmitter, aligned in its direction. In the NLOS scenario (green line), the transmitter and receiver once again face each other but are placed at the outer walls of rooms 5 and 1, respectively. The alignment of antennas is achieved by fine-adjusting the receiver's horizontal position in the room based on the RSSI at the receiver's end. As in the LOS scenario, activity images are captured in the NLOS scenario using an ESP32-S3-based camera board placed in the room where the activity occurs. 

\subsection{Signal Strength}
\begin{figure}[t!] 
  \centering
  \includegraphics[width=\linewidth]{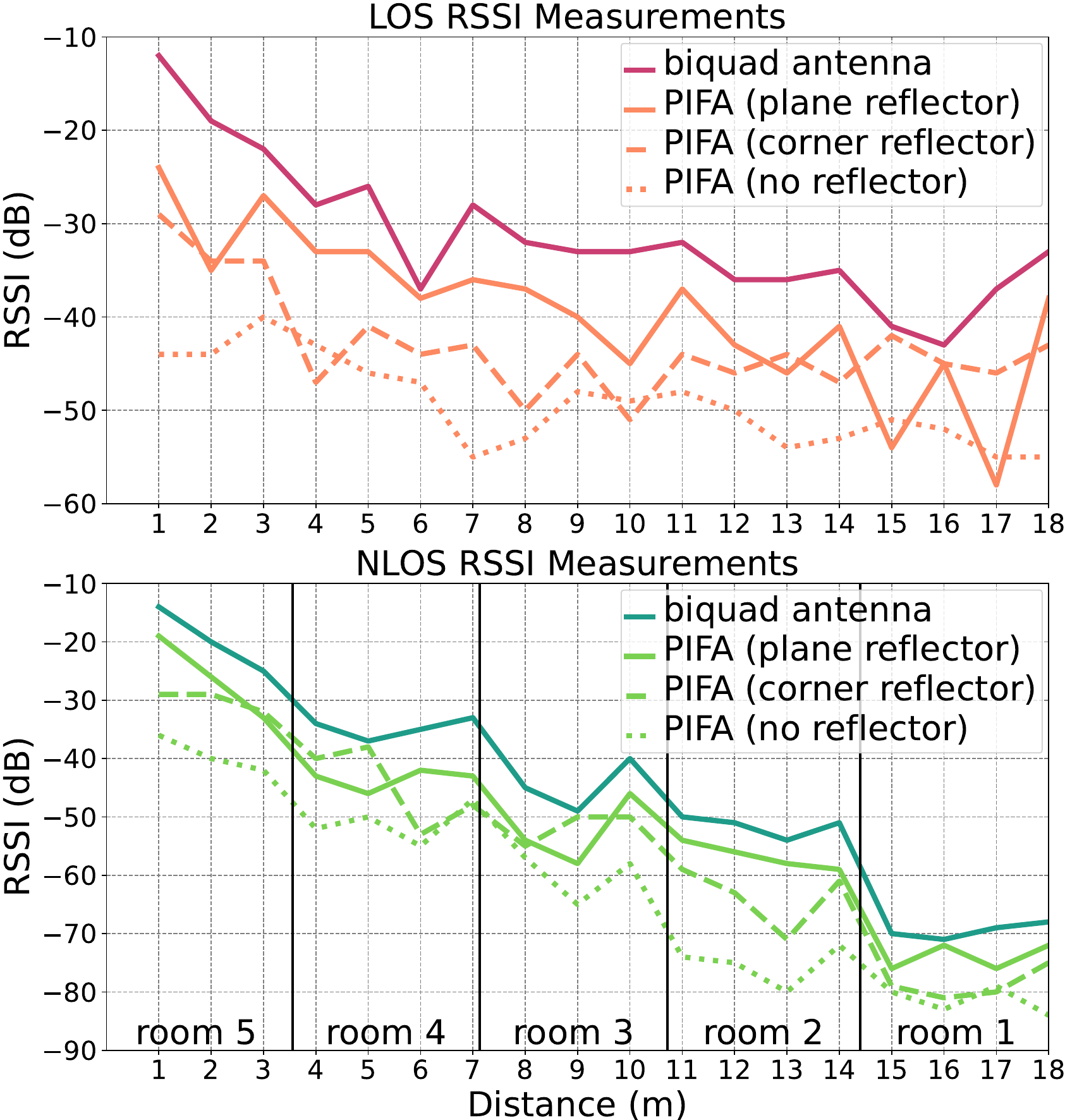}
  \caption{Comparison of LOS and NLOS signal strength (RSSI) between systems over a distance of 18m and five rooms.}
  \label{fig:rssi}
  \vspace{-2mm}
\end{figure}

To identify candidate systems for long-range through-wall HAR applications, a signal strength evaluation based on the RSSI in LOS and NLOS scenarios is conducted. For this purpose, a transmitter-receiver pair of each system is deployed in the evaluation environment. Starting with a transmitter-receiver spacing of 1m, the receiver is moved away from the transmitter in increments of 1m, up to a maximum distance of 18m. At each position, we measure the corresponding signal strength by computing the mean RSSI of 1k CSI packets. The results of this experiment are visualized in Figure \ref{fig:rssi}, showing the RSSI measurements of all systems in both LOS and NLOS scenarios.

Focusing on the LOS scenario, it can be observed that the biquad antenna system consistently outperforms all other systems across the tested range. Furthermore, the signal strength of PIFA-based systems is significantly enhanced by adding reflectors. Both the plane reflector and 90$^{\circ}$ corner reflector systems exhibit improved signal strength compared to the baseline system. Interestingly, the plane reflector outperforms the 90$^{\circ}$ corner reflector despite its relative simplicity. We suspect this might be due to destructive interference caused by the inconsistent spacing between the PIFA and the reflector. While the spacing is 15mm (1/8-wavelength) at the center of the PIFA, it decreases as we move along the horizontal direction due to the reflector's geometry.

In the NLOS scenario, we observe a similar trend, with the biquad antenna system consistently outperforming other systems. However, the signal strength differences between systems are less pronounced. Furthermore, as expected, the reduction in signal strength with respect to distance is more significant in the NLOS scenario. For the biquad and PIFA with plane reflector systems, although still adequate for a stable connection, the RSSI at a distance of 18m drops from -33dB to -68dB ($\downarrow$35dB) and from -38dB to -72dB ($\downarrow$34dB), respectively. In comparison, at the same distance, the RSSI of the baseline system drops from -55dB to -84dB ($\downarrow$29dB), leading to frequent packet loss and an unstable connection. While the baseline system achieves sufficient signal strength in the LOS scenario, it might not be well-suited for long-range NLOS scenarios. Based on these results, the most promising candidates for long-range through-wall HAR are the biquad antenna and the PIFA with plane reflector systems, which are further evaluated in the remainder of this work.

\subsection{Data}
\label{subsec:data}
To assess LOS and NLOS HAR performance of the biquad and PIFA with plane reflector systems, we collect the Wallhack1.8k dataset, comprising 1806 CSI amplitude spectrograms of human activities collected in the evaluation environment, which is used for the training of CNN-based HAR models. The objective is to distinguish between coarse and fine body movements (e.g., walking vs. arm movements). For data collection, activities are conducted within five circular activity zones (1.5m radius) along the LOS and NLOS transmission paths, as shown in Figure \ref{environment}. These zones are located at distances of $\{1.8, 5.4, 9.4, 13.0, 16.6\}$m from the receiver, corresponding to room centers in the NLOS scenario. For both systems, we record two minutes of continuous \textit{walking} and \textit{walking + arm-waving} activities in each activity zone, as well as five minutes of \textit{no presence} (no person in the recording environment) for each scenario.

\begin{table}[t!]
\caption{Distribution of samples across subsets of the Wallhack1.8k dataset. $^{*}$PIFA samples were collected using the plane reflector system depicted in Figure \ref{fig:antennasPCBPlane}.}
\centering
\begin{tabularx}{0.48\textwidth}{>{\arraybackslash}p{11mm}>{\centering\arraybackslash}p{12mm}>{\centering\arraybackslash}p{12mm}>{\centering\arraybackslash}p{9mm}>{\centering\arraybackslash}p{8mm}Y}
\toprule
Subset & Scenario & Antenna & Rooms & Classes & Samples\\
\midrule
W1.8k$_{LB}$ & LOS & biquad & 1 & 3  & 458\\
W1.8k$_{LP}$ & LOS & PIFA$^{*}$  & 1 & 3  & 461\\
W1.8k$_{NB}$ & NLOS & biquad  & 5 & 3 & 450\\
W1.8k$_{NP}$ & NLOS & PIFA$^{*}$ & 5 & 3 & 437\\
\midrule    
 &  &  &   & Total: & 1806\\
\bottomrule
\end{tabularx}
\label{wallhackOverview}
\vspace{-5mm}
\end{table}

\begin{figure}[t!]
  \centering
  \includegraphics[width=\linewidth]{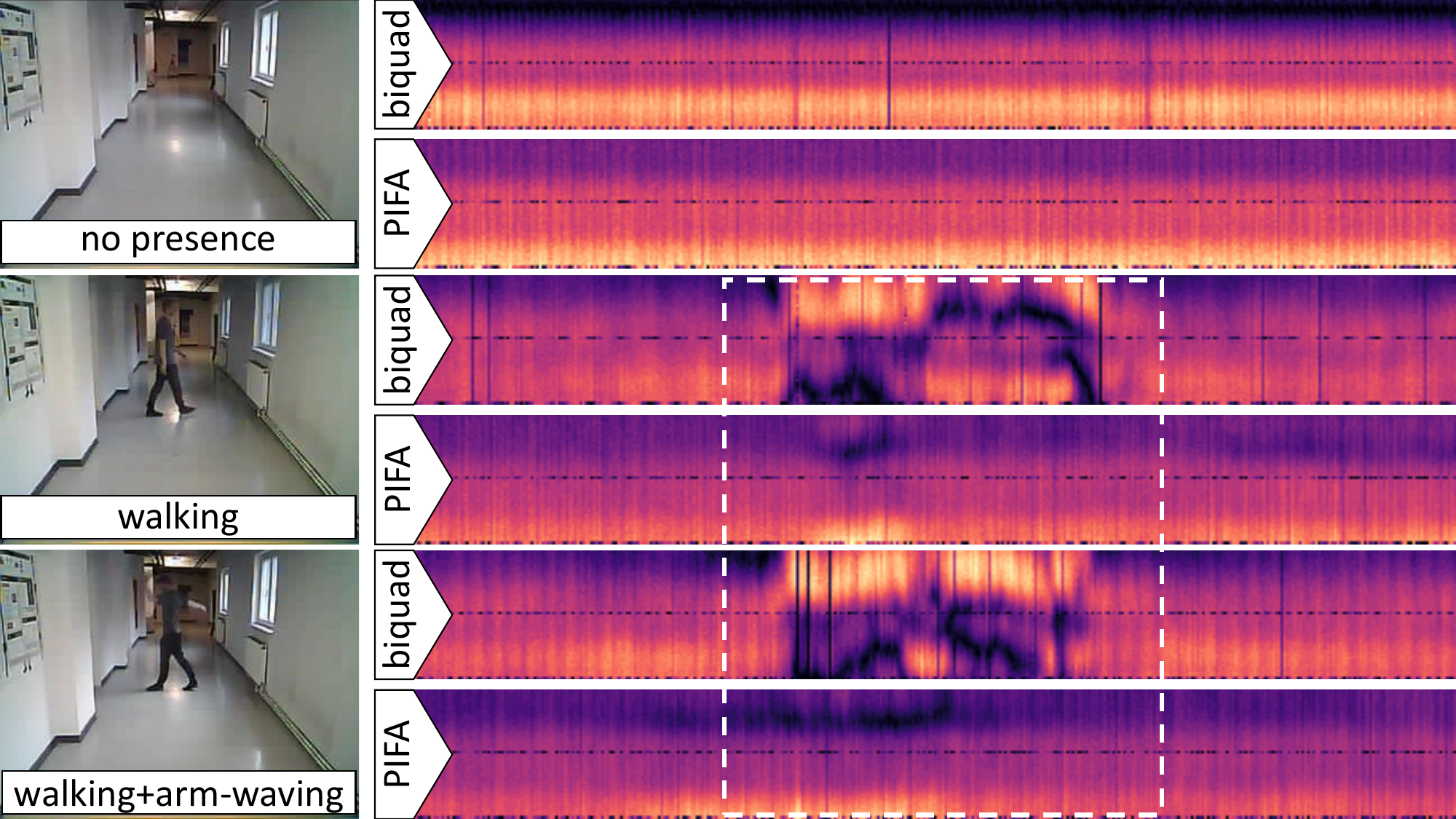}
  \caption{LOS CSI amplitude spectrograms of the classes \textit{no presence}, \textit{walking}, and \textit{walking + arm-waving}, captured with biquad antenna and PIFA with plane reflector systems at a distance of 9.4m (room 3 in the NLOS scenario). The spectrograms show the amplitudes of 52 L-LTF subcarriers over a time interval of 4 seconds ($\sim$400 packets).}
  \label{fig:activities}
  \vspace{-5mm}
\end{figure}

As a pre-processing step, the recorded CSI time series data is trimmed using the corresponding RGB images to remove any CSI packets that do not correspond to the target activity. Additionally, we perform outlier removal using the Hampel filter \cite{Pearson2016}. The filtered CSI time series data is then transformed into time-frequency plots of subcarrier amplitudes over time. To achieve this, the filtered CSI time series data is divided into segments of 400 CSI packets (equivalent to 4-second time intervals at a sending frequency of 100Hz), and the amplitudes of 52 Legacy Long Training Field (L-LTF) subcarriers are plotted, resulting in a spectrogram size of 400$\times$52. LOS spectrogram examples of all activities are given in Figure \ref{fig:activities}. For the training of activity recognition models, we assign the labels $\{0,1,2\}$, corresponding to the classes \textit{no presence}, \textit{walking}, and \textit{walking + arm-waving}. This data collection process for both systems in both LOS and NLOS scenarios leads to four subsets constituting the Wallhack1.8k dataset. We adopt the subset naming convention W1.8k$_{XY}$, where X represents the scenario (L=LOS or N=NLOS) and Y represents the system (B=biquad or P=PIFA). The distribution of the 1806 spectrograms in Wallhack1.8k across subsets is given in Table \ref{wallhackOverview}. 

Recognizing the potential of Wallhack1.8k as a benchmark for evaluating HAR performance in long-range LOS and NLOS scenarios, along with assessing model generalization across scenarios and systems, an open problem in WiFi-based HAR \cite{Chen2023}, the collected data, including raw CSI recordings, spectrograms, and labels, is made publicly available\footnote{Wallhack1.8k, \href{https://zenodo.org/record/8188999}{https://zenodo.org/record/8188999}}.

\section{Evaluation}
\label{sec:evaluation}
As demonstrated in \cite{Gao2017}, CSI spectrograms can be efficiently processed using CNNs to enable a variety of HAR applications. Building on this approach, we train HAR models on the four subsets of the Wallhack1.8k dataset to measure and compare system performance in both LOS and NLOS scenarios. 

\begin{figure}[t!]
  \centering
  \includegraphics[width=\linewidth]{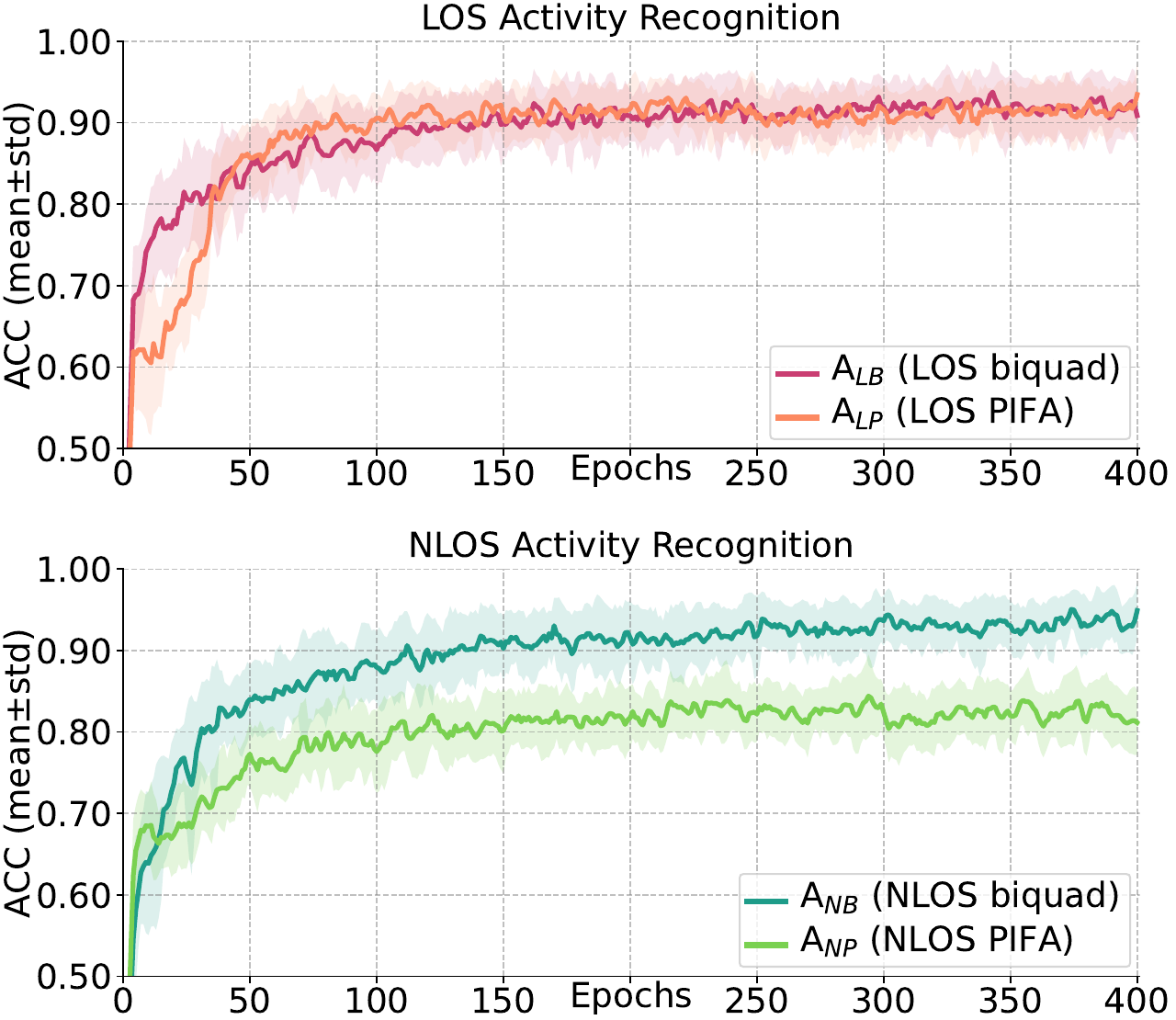}
  \caption{Validation accuracy (mean$\pm$std) of activity recognition models, measured across ten independent training runs spanning 400 epochs.}
  \label{fig:ARTraining}
  \vspace{-1mm}
\end{figure}

\subsection{Model Training}
To ensure the reproducibility of our results, baseline HAR models are trained on Wallhack1.8k using the standard implementation of the EfficientNetV2 small architecture from \emph{torchvision.models}\footnote{EfficientnetV2s, \href{https://pytorch.org/vision/main/models/efficientnetv2.html}{https://pytorch.org}, accessed: 10-16-2023} \cite{tan2021efficientnetv2}. EfficientNetV2 small is a lightweight feature extractor commonly employed as a backbone. The resulting activity recognition models (A$_{LB}$, A$_{LP}$, A$_{NB}$, and A$_{NP}$) follow the naming convention of the Wallhack1.8k subsets, as given in Table \ref{wallhackOverview}. The suffix indicates the subset on which a model is trained. For training, each subset is divided into training, validation, and test sets using an 8:1:1 split ratio. All models are trained from scratch to eliminate any prior knowledge derived from pre-training on RGB images that could influence the results. The models undergo 400 epochs of training using the Adam optimizer with a learning rate of 0.0001 and a batch size of 16. Additionally, a balanced sampler is used to mitigate class imbalances in the training dataset. As data augmentation, only random circular rotations along the time axis are applied to spectrograms. For each system and scenario, ten independent training runs are conducted and the respective model instance with the highest validation performance is selected as the final model. We then compute and report the mean and standard deviation of metrics such as precision, recall, F1-score, and classification accuracy of these models on the respective test dataset. The training progress of all models is visualized in Figure \ref{fig:ARTraining}.

\begin{table}[t!]
\caption{LOS and NLOS activity recognition results for the biquad antenna and PIFA with plane reflector systems, measured on the test subsets of Wallhack1.8k.}
\label{tab:activityRecognitionResultsLOS}
\centering
\begin{tabularx}{0.48\textwidth}{>{\arraybackslash}p{5mm}>{\centering\arraybackslash}p{12mm}>{\centering\arraybackslash}p{12mm}>{\centering\arraybackslash}p{12mm}>{\centering\arraybackslash}p{12mm}Y}
\toprule
Model & Test & Precision & Recall & F1 & ACC\\
\midrule
A$_{LB}$ & W1.8k$_{LB}$  & 90.0$\pm$2.5 & 89.0$\pm$2.2 & 89.5$\pm$2.3 & 89.4$\pm$2.3 \\
A$_{LP}$ & W1.8k$_{LP}$  & 91.1$\pm$3.9 & 90.4$\pm$4.0 & 90.8$\pm$4.0 & 90.2$\pm$4.3 \\
\midrule
A$_{NB}$ & W1.8k$_{NB}$  & 92.8$\pm$3.3 & 91.2$\pm$3.9 & 92.0$\pm$3.5 & 92.0$\pm$3.5 \\
A$_{NP}$ & W1.8k$_{NP}$  & 86.6$\pm$5.0 & 86.2$\pm$4.9 & 86.4$\pm$4.9 & 86.8$\pm$4.7 \\
\bottomrule
\end{tabularx}
\vspace{-5mm}
\end{table}

\subsection{Activity Recognition Results}
The activity recognition results for LOS (A$_{LB}$ and A$_{LP}$) and NLOS scenarios (A$_{NB}$ and A$_{NP}$) are given in Table \ref{tab:activityRecognitionResultsLOS}. In the LOS scenario, both systems achieve comparable recognition accuracies, with A$_{LP}$ slightly outperforming A$_{LB}$ (90.2$\pm$4.3 vs. 89.4$\pm$2.3). This result is surprising, as the amplitude patterns in spectrograms captured by the biquad antenna system are highly pronounced, while they are barely noticeable in spectrograms captured by the system using the PIFA with a plane reflector (see Figure \ref{fig:activities}). We hypothesize that while activity spectrograms captured by the biquad antenna system seemingly contain more problem-relevant information, the relative differences in amplitude might be more significant due to the higher sensitivity of the biquad antenna, while the underlying pattern is the same with both systems. Consequently, LOS spectrograms captured by the biquad antenna system, despite showing more pronounced patterns, would not contain additional information from which a model could benefit.

While in the LOS scenario, both systems achieve comparable performance with models exhibiting similar training behavior, a deviation from this trend is noticeable in the training progress of NLOS models, as visualized in Figure \ref{fig:ARTraining}. Early in the training, the validation accuracies of A$_{NB}$ and A$_{NP}$ diverge, creating a consistent gap that persists throughout training. As a result, the NLOS performance metrics, provided in Table \ref{tab:activityRecognitionResultsLOS}, reveal a significant difference between the systems. A$_{NB}$ outperforms A$_{NP}$ in recognition accuracy by 5.2 percentage points (92.0$\pm$3.5 vs. 86.8$\pm$4.7), demonstrating the superiority of the biquad antenna system in the core NLOS HAR scenario.

\section{Conclusion}
\label{sec:conclusion}
In this work, we evaluated four ESP32-S3-based systems with different antenna configurations for long-range through-wall HAR. A biquad antenna-based system and a system combining the ESP32-S3's PIFA with a plane reflector exhibited superior LOS and NLOS signal strength when deployed in a challenging office environment spanning five rooms (18m). We created the Wallhack1.8k dataset, comprising 1806 CSI amplitude spectrograms of human activities. Wallhack1.8k is made publicly available and is intended as a benchmark for developing methodologies for WiFi CSI-based HAR, including cross-scenario and cross-system generalization techniques. Using Wallhack1.8k, we trained baseline LOS and NLOS activity recognition models with EfficientNetV2 architecture, demonstrating the feasibility of long-range through-wall HAR with the proposed systems. Notably, the biquad antenna system achieved the highest accuracy for the core NLOS activity recognition problem at 92.0$\pm$3.5, surpassing the PIFA-based system by 5.2 percentage points (86.8$\pm$4.7)




\bibliographystyle{IEEEtran}
\bibliography{IEEEexample}

\end{document}